\documentclass[conference]{IEEEtran}
\usepackage{blindtext, graphicx,booktabs}
\usepackage[top=1in,bottom=0.75in,right=0.75in,left=0.75in]{geometry}
\usepackage{amsmath}

\ifCLASSINFOpdf
\else
\fi
\begin{document}
%
\title{Enhancement of Low-cost GNSS Localization in Connected Vehicle Networks Using Rao-Blackwellized Particle Filters}

\author{\IEEEauthorblockN{ }
\IEEEauthorblockA{Macheng Shen \\Department of Naval Architecture \\
 and Marine Engineering\\
University of Michigan \\
Ann Arbor, Michigan 48109\\
Email: macshen@umich.edu }
\and
\IEEEauthorblockN{ }
\IEEEauthorblockA{Ding Zhao \\
 Transportation Research Institute\\
 University of Michigan\\
 Ann Arbor, Michigan 48109\\
 Email: zhaoding@umich.edu
 }
\and
\IEEEauthorblockN{ \\  }
\IEEEauthorblockA{ Jing Sun\\
Department of Naval Architecture \\
 and Marine Engineering\\
University of Michigan \\
Ann Arbor, Michigan 48109\\
Email: jingsun@umich.edu
 }}


%


\maketitle

\begin{abstract}
An essential function for automated vehicle technologies is accurate localization. It is  difficult, however, to achieve lane-level accuracy with low-cost Global Navigation Satellite System (GNSS) receivers due to the biased noisy pseudo-range measurements. Approaches such as Differential GNSS can improve the accuracy, but usually require an enormous amount of investment in base stations. The emerging connected vehicle technologies provide an alternative approach to improving the localization accuracy. It has been shown in this paper that localization accuracy can be enhanced by fusing GNSS information within a group of connected vehicles and matching the configuration of the group to a digital map to eliminate the common bias in localization. A Rao-Blackwellized particle filter (RBPF) was used to jointly estimate the common biases of the pseudo-ranges and the vehicles positions. Multipath biases, which are non-common to vehicles, were mitigated by a multi-hypothesis detection-rejection approach. The temporal correlation was exploited through the prediction-update process. The proposed approach was compared to the existing static and smoothed static methods in the intersection scenario. Simulation results show that the proposed algorithm reduced the estimation error by fifty percent and reduced the estimation variance by two orders of magnitude.
\end{abstract}


%
\IEEEpeerreviewmaketitle

\section{Introduction}
An essential function of intelligent transportation systems is accurate localization. One common solution to the vehicle localization problem is the Global Navigation Satellite System (GNSS) which calculates the absolute position of the vehicle from pseudo-range measurements. Pseudo-range measurements are different from the actual distances between the receiver and the satellites. This difference results in localization inaccuracy, which can be decomposed into common errors due to satellite clock error, ionospheric and tropospheric delay and non-common errors due to receiver noise and multipath biases. The nominal accuracy for a single-band receiver is about 10 to 20 meters, which results in a position error of several meters \cite{r3}. Thus, it is very difficult to achieve lane-level accuracy robustly using low-cost GNSS receivers alone.\\
\indent Differential GNSS (DGNSS) is an enhancement to GNSS that improves the localization accuracy to the sub-meter level through correction of the common biases broadcasted by a network of fixed reference stations. Centimeter-level accuracy is achievable by the Real Time Kinematic (RTK) technique, which uses carrier phase measurements to provide real-time corrections \cite{r4}. These techniques, however, rely on infrastructure that is expensive and not readily available. It would be highly desirable, therefore, to develop low-cost solutions for lane-level accurate localization.
\section{Related Works}
The accuracy of GNSS localization can also be improved using either sensor fusion or map matching, with some promising results shown in recent years \cite{r5,r6,r7}. \\
\indent Another low-cost navigation solution is the Inertial Navigation System (INS). The problem, however, is that due to sensor noise, the navigation error using INS alone can grow without bounds. In contrast, GNSS provides an absolute position measurement whose error is always bounded. Data fusion algorithms such as Extended Kalman Filter (EKF) \cite{r8}, Unscented Kalman Filter (UKF) \cite{r9} and particle filter \cite{r10} can be used to fuse the complementary measurements from low-cost INS and GNSS sensors to produce high accuracy navigation solutions.\\
\indent With the rapid development of digital maps, navigation algorithms based on map matching have also been extensively studied \cite{r6,r7,r11,r12}. Map matching algorithms match the noisy raw GNSS positioning results to a trajectory that is compatible with road constraints. Additional sensors such as cameras allow for accurate localization down to the centimeter-level \cite{r12}. Nevertheless, these approaches all require infrastructure or additional sensors to reach lane-level accuracy. Wide application is unlikely in the near future. Therefore, the ability to use low-cost GNSS receivers alone while achieving the same degree of accuracy is highly desired. \\
\indent Recently, Dedicated Short Range Communications (DSRC)--based V2V technique has been developed and validated in the real world \cite{r21}. The introduction of connected vehicles provides an alternative for improving navigation solutions through V2X, such that map matching can be conducted cooperatively to correct the common localization error \cite{r13}. Assuming that most vehicles travel within lanes, the correction to the common localization biases can be determined so that the corrected positions of a group of vehicles best fit the road map. This requires communication within the group of vehicles, hereafter referred to as cooperative map matching (CMM). Schubert et al.\cite{r14} have shown the advantage of exchanging raw GNSS measurements and vehicle relative distances have been shown. More recently, the capability of CMM to mitigate the biases from the localization results obtained from low-cost GNSS receivers alone has been demonstrated in Rohani et al.\cite{r1}. \\
\indent Three main difficulties arise for CMM using low-cost GNSS receivers alone are:
\begin{itemize}
 \item Effect of non-common error: In the presence of non-common errors, a correction for the common localization error that makes all vehicle positions compatible with the road constraints, may not exist. If the road constraints are enforced aggressively, without consideration for the non-common errors, the localization solution may overly converge, that is, the variance of the estimation may be underestimated. 
 \item Correlation between common biases and vehicles positions estimation: An estimation error of the common biases may induce an estimation error of the vehicles positions and vice versa. If they are estimated sequentially and the respective estimation results are then fused with each other for a more comprehensive estimation, the corresponding data fusion scheme must be carefully selected to avoid over-convergence due to fusion of correlated data. 
 \item Application of road constraints: Road constraints may be regarded as additional observations to correct for the estimation of vehicles positions. However, unlike sensor observations, road constraints cannot be described analytically by observation equations, which makes it almost impossible for traditional filtering schemes based on the Kalman filter to implement the map matching. 
 \end{itemize}
\indent
\indent Rohani et al. \cite{r1} presented a particle-based CMM algorithm to address these three difficulties. The first difficulty, that of the effect of non-common pseudo-range error was considered by a weighted road map approach to preserve consistency. The second difficulty is handled by tracking the origin of the common bias corrections from different vehicles and fusing only those corrections from independent sources to avoid data incest, which effectively avoids over-convergence, though some of the correlated corrections containing additional information have to be discarded. The third difficulty is handled by a particle-based method that uses only the vehicle positions' estimation of the current epoch. Algorithms that better utilizes all available data are expected to yield improved localization performance.\\
\indent In this work, the problem of inferring the true vehicle positions, as well as the GNSS common biases from the biased noisy pseudo-range observables of a group of vehicles given a precise digital map, is addressed by a Bayesian filtering approach. The aforementioned difficulties are solved by jointly estimating the common biases and vehicle positions using a Rao-Blackwellized particle filter (RBPF). In the RBPF, the correlation between common biases and vehicle positions is modeled implicitly through the diversity of the particles. As a result, there is no need for explicit data fusion. The effect of the multipath biases is mitigated through a detection-rejection method based on a statistical test. The particle filter structure allows multiple hypotheses with respect to the detection of multipath biases, thus making the detection more robust. In addition, the particle filter is flexible enough to handle road constraints in a straightforward manner by manipulating the particle weights according to the road constraints. It also fully exploits the temporal correlation through a prediction-update process, eliminating impossible configurations in the joint space of common biases and vehicle state variables, drastically reducing the estimation variance. The computational complexity only varies linearly with respect to the number of vehicles, which makes the proposed RBPF both effective and efficient.\\
\indent In the following sections, the formulation and implementation of the RBPF for CMM are presented. In Section 2, the derivation of the RBPF is presented in detail, and the localization enhancement algorithm is described with the assumptions under which the algorithm is valid. Simulation results are presented in Section 3. The simulation scenario and noise models are introduced, followed by the performance analysis of the RBPF as it is compared with those of the existing methods. Conclusions are presented in Section 4.
\section{Theory and Method}
In this section, the theoretical aspect of the RBPF is introduced first. Then the prediction-update structure of the RBPF is shown in detail. \\
\indent Particle filter takes the Bayes filter algorithm and uses a set of particles to represent the posterior probability distribution of the estimated quantities conditioned on the measurements. In the prediction step, each particle predicts its state according to the process model; in the update step, the weight of each particle is calculated according to the principle of importance sampling given the measurements; then the particles are resampled according to the weight to avoid weight degeneracy \cite{r15}. Although the particle filter is computationally intractable if the dimension of the state vector is high, the Rao-Blackwellization technique takes advantage of the conditional independence property so that it is only necessary to represent a subspace of the state vector using particle filters, conditioned on which the remaining independent subspace can be marginalized out analytically \cite{r16}.\\ 
\indent The following assumptions are made in this work:
\begin{enumerate}
 \item The non-common errors of different vehicles are uncorrelated.
 \item The common biases vary slowly over time.
 \item The vertical positions of the vehicles can be obtained with reasonable accuracy from the digital map.
 \end{enumerate}

\indent Assuptions (1) - (3) are unrestrictive in the following senses. The first assumption is valid as long as the participating vehicles are not concentrated in the same area; otherwise, the multipath errors may be correlated. It will be true for most rural areas and some urban areas. The second assumption is reasonable because the tropospheric and ionospheric delays, as the major components of the common biases, typically change very slowly over time \cite{r3}. The third assumption is also reasonable, as the difference between the vertical positions recorded by the digital map and the ground truth can be considered equivalent measurement noise and accounted for by increasing the noise variance parameter.\\
\indent The vehicle cooperative localization has been formulated in this work as a Bayesian filtering problem which estimates the joint posterior distribution of the GNSS common biases and the states of the vehicles conditioned on the raw pseudo-range measurements. Assumption 1 implies that, conditioned on the common biases, the posterior distributions of the vehicle states are independent from each other, which means that the joint distribution can be factorized as follows:
\begin{equation}
\begin{aligned}
p(C^{1:Ns}_{1:t},X^{1:Nv}_{1:t}|Z_{1:t})=p(X^{1:Nv}_{1:t}|C^{1:Ns}_{1:t},Z_{1:t})p(C^{1:Ns}_{1:t}|Z_{1:t})\\=\prod^{Nv}_{i=1}p(X^{i}_{1:t}|C^{1:Ns}_{1:t},Z_{1:t})p(C^{1:Ns}_{1:t}|Z_{1:t}),
\end{aligned}
\end{equation}
where $C^{1:Ns}_{1:t}$ denotes the common biases of the total $Ns$ visible satellites, $X^{1:Nv}_{1:t}$ denotes the state vectors of the total $Nv$ vehicles, and $Z_{1:t}$ denotes all the pseudo range measurements, from time 1 to time $t$.\\
\indent The posterior distribution of the common biases $p(C^{1:Ns}_{1:t}|Z_{1:t})$ is estimated by a particle filter, and the conditionally independent posterior distribution of the vehicle states $p(X^{i}_{1:t}|C^{1:Ns}_{1:t},Z_{1:t})$ are estimated by independent EKFs.
\subsection{Prediction of States}
Assumption 2 implies that the common biases can be modeled by the first-order Gaussian-Markov process:
\begin{equation}
C^j_t=C^j_{t-1}+w^j_t{\Delta}t,   
\end{equation}
where $w^j_t{\sim}N(0,{\sigma}^2_c)$, with ${\sigma}^2_c$ denoting the variance of the common bias drift, ${\Delta}t$ is the length of the time interval between two successive updates of the states and $j=1,2,...,Ns$ is the index for satellites. \\
\indent Assumption 3 implies that only the horizontal positions and velocities need be modeled explicitly. Therefore, the state vector of the $i$th vehicle is\\
\begin{equation}
X^i_t=(\begin{array}{cccccc}
     x^i_t&{\dot x}^i_t&y^i_t&{\dot y}^i_t&b^i_t&{\dot b}^i_t\\
\end{array})^T,
\end{equation}
where $x^i_t$ and $y^i_t$ are the horizontal positions; ${\dot x}^i_t$ and ${\dot y}^i_t$ are the horizontal velocities; and $b^i_t$ and ${\dot b}^i_t$ are the receiver clock bias and drift, respectively.\\
\indent The mean is propagated by\\
\begin{equation}
\bar{X}^i_t=AX^i_{t-1},\textrm{with }
A=\left[
\begin{matrix}
B&0&0\\
0&B&0\\
0&0&B
\end{matrix}
\right],B=\left[
\begin{matrix}
1&{\Delta}t\\
0&1
\end{matrix}
\right],
\end{equation}

where $\bar{X}^i_t$ is the predicted mean.\\
\indent The associated covariance matrix is propagated by\\
\begin{equation}
\begin{aligned}
&\bar{\Sigma}^i_t=A{\Sigma}^i_{t-1}A^T+R_t,\textrm{with }
R_t=\left[
\begin{matrix}
R_x&0&0\\
0&R_y&0\\
0&0&R_b
\end{matrix}
\right],\\
&R_x=\left[
\begin{matrix}
\frac{\sigma^2_{ax}{\Delta}t^4}{4}&\frac{\sigma^2_{ax}{\Delta}t^3}{2}\\
\frac{\sigma^2_{ax}{\Delta}t^3}{2}&\sigma^2_{ax}{\Delta}t^2
\end{matrix}
\right],
R_y=\left[
\begin{matrix}
\frac{\sigma^2_{ay}{\Delta}t^4}{4}&\frac{\sigma^2_{ay}{\Delta}t^3}{2}\\
\frac{\sigma^2_{ay}{\Delta}t^3}{2}&\sigma^2_{ay}{\Delta}t^2
\end{matrix}
\right],\\
&R_b=\left[
\begin{matrix}
\frac{\sigma^2_{d}{\Delta}t^4}{4}+\sigma^2_{b}{\Delta}t^2&\frac{\sigma^2_{d}{\Delta}t^3}{2}\\
\frac{\sigma^2_{d}{\Delta}t^3}{2}&\sigma^2_{d}{\Delta}t^2
\end{matrix}
\right],
\end{aligned}
\end{equation}
where ${\Sigma}^i_{t-1}$ is the covariance matrix of the state vector; $\bar{\Sigma}^i_t$ is the predicted covariance matrix; $\sigma^2_{ax}$ and $\sigma^2_{ay}$ are the variances of the horizontal accelerations; and $\sigma^2_{b}$ and $\sigma^2_{d}$ are the variances of the clock bias and drift time derivatives \cite{r17}, which are assumed to be uncorrelated during the derivation.
\subsection{Multipath Rejection and Measurement Update}
The pseudo-range measurement model between satellite $j$ and vehicle $i$ is\\
\begin{equation}
Z^{j,i}_t={\Vert}p^i_t-s^j_t{\Vert}+C^j_t+b^i_t+\lambda^{j,i}_tm^i_t+v^i_t,
\end{equation}
where $p^i_t$ is the position of the vehicle and $s^j_t$ is the satellite position; $\lambda^{j,i}_t$ is a binary indicator variable. $m^i_t$ is the potential multipath bias and $v^i_t{\sim}N(0,{\sigma}^2_z)$ is the receiver noise, which is assumed to be white. The binary value of the indicator variable is to be determined through a $\chi^2$ test to indicate the presence of multipath bias.\\
\indent In the absence of multipath biases, the predicted mean of the pseudo-range measurement will be
\begin{equation}
\bar{Z}^{j,i}_t={\Vert}p^i_t-s^j_t{\Vert}+C^j_t+b^i_t
\end{equation}
\indent The difference between the actual pseudo-range measurement and the predicted mean will obey a Gaussian distribution, and the Mahalanobis distance of this random variable will obey the $\chi^2$ distribution of one degree of freedom:\\
\begin{equation}
D^2_{j,i}=(Z^{j,i}_t-\bar{Z}^{j,i}_t)^TP^{-1}_{j,i}(Z^{j,i}_t-\bar{Z}^{j,i}_t){\sim}\chi^2_1
\end{equation}
\begin{equation}
P_{j,i}=H_{j,i}\Sigma^i_{xy}H^T_{j,i}+\sigma^2_z, H_{j,i}=\frac{{\partial}Z^{j,i}_t}{\partial(x^i_t,y^i_t)},
\end{equation}
where $\Sigma^i_{xy}$ is the submatrix of the covariance matrix representing the uncertainty of the horizontal position, and $H_{j,i}$ is the Jacobian of the measurement function with respect to the horizontal position, which projects the uncertainty of the position space to the range space.\\
\indent The indicator variable is determined by\\
\begin{equation}
\lambda^{j,i}_t=\left\{
 \begin{array}{rcl}
 0       &      & D^2_{j,i}{\leq}F^{-1}(\alpha_1)\\
 1       &      & D^2_{j,i}{\geq}F^{-1}(\alpha_2){\Vert}u_{j,i}{\leq}\frac{F(D^2_{j,i})-\alpha_1}{\alpha_2-\alpha_1}
 \end{array} \right.,
\end{equation}
where $F$ is the Cumulative Distribution Function (CDF) of the $\chi^2_1$ distribution. $\alpha_1$ , $alpha_2$ (with $\alpha_1<\alpha_2$) are the confidence levels for the rejection and acceptance of the multipath presence hypothesis, respectively. $u_{j,i}$ is a random number generated according to the uniform distribution on $[0,1]$, $\Vert$ is the logical "or".\\
\indent $\alpha_1$ is chosen as a small number such that only those measurements compatible with the predicted measurements are accepted with certainty as normal, while $\alpha_2$ is chosen as a large number such that only those measurements incompatible with the predicted measurement are rejected with certainty as spurious. In the intermediate cases, the measurements are accepted with probability. As a result, the particle filter keeps multiple hypotheses with respect to the assumptions on multipath biases. Particles that make wrong hypotheses will be eliminated by applying the map constraints.\\
\indent The weights of the particles are calculated according to the principal of importance weight. The detailed mathematical derivation can be found in \cite{r18}. For the pseudo-range measurement from the $j$th satellite, the weights of the particles are updated as follows\\
\begin{equation}
w^{[k]}_j=\left\{
 \begin{array}{rcl}
 w^{[k]}_{j-1}\frac{1}{2\pi P_{j,i}}exp(-\frac{1}{2}D^2_{j,i})       &      & \lambda^{j,i}_t=0\\
 w^{[k]}_{j-1}\frac{1}{2\pi P_{j,i}}exp(-\frac{1}{2}F^{-1}(\alpha_3))       &      & \lambda^{j,i}_t=1
 \end{array} \right.,
\end{equation}
where $\alpha_3$ is a parameter that can be tuned depending on the environment and frequency of multipath occurrence, and superscript $[k]$ is the index of particles.\\
\indent The vehicle states are then updated using all the pseudo-range measurements regarded as free of multipath biases, that is, with $\lambda^{j_n,i}_t=0, n=1,2...N$ by
\begin{equation}
X^i_t=\bar{X}^i_t+K^i_t(Z^i_t-\bar{Z}^i_t),
\end{equation}
where $\bar{X}^i_t$ and $\bar{Z}^i_t$ are calculated by Eq. (5) and Eq. (14), respectively. $Z^i_t$ is the actual measurement. The Kalman filter matrix $K^i_t$ is calculated as:
\begin{equation}
K^i_t=\bar{\Sigma}^i_t(\tilde{H}^i_t)^T(\tilde{H}^i_t\bar{\Sigma}^i_t\tilde{H}^i_t)^T+Q_t)
\end{equation}
\begin{equation}
\Sigma^i_t=(I-K^i_t\tilde{H}^i_t)\bar{\Sigma}^i_t
\end{equation}
\begin{equation}
\tilde{H}^i_t=\left[
\begin{matrix}
\frac{{\partial}Z^{j_1,i}}{{\partial}x^i_t}&0&\frac{{\partial}Z^{j_1,i}}{{\partial}y^i_t}&0&0&0\\
...&...&...&...&...&...\\
\frac{{\partial}Z^{j_N,i}}{{\partial}x^i_t}&0&\frac{{\partial}Z^{j_N,i}}{{\partial}y^i_t}&0&0&0
\end{matrix}
\right]_{N{\times}6}
\end{equation}
\begin{equation}
Q_t=\left[
\begin{matrix}
\sigma^2_z&0&0\\
...&...&...\\
0&0&\sigma^2_z
\end{matrix}
\right]_{N{\times}N},
\end{equation}
where $\bar{\Sigma}^i_t$ is calculated by Eq. (8); $\tilde{H}^i_t$ is the measurement Jacobian for batch update; $I$ is the identity matrix.
\subsection{Applying Map Constraint}
After the conventional EKF update and weight determination, the map constraint is used to further modify the particle weights such that those particles with vehicle configurations incompatible with the map constraint are assigned a low weight and will be eliminated with high probability during the resampling. In this paper, the particle weights are modified by\\
\begin{equation}
\begin{aligned}
w^{[k]}_i=w^{[k]}_{i-1}{\int}\varepsilon(x^i_t,y^i_t)p(x^i_t,y^i_t)dx^i_tdy^i_t, i=1,2...Nv,\\
\textrm{with }
\varepsilon(x^i_t,y^i_t)=\left\{
 \begin{array}{rcl}
 1       &      & (x^i_t,y^i_t) \mbox{on lane}\\
 0       &      & (x^i_t,y^i_t) \mbox{out of lane}
 \end{array} \right.,
 \end{aligned}
\end{equation}
where $p(x^i_t,y^i_t)$ is the joint Gaussian distribution drawn from the EKF.\\
\indent The integral in Eq. (24) is difficult to calculate analytically due to the potentially complicated geometry. Therefore, it is again calculated by Monte Carlo Integration, where the proposal distribution is $p(x^i_t,y^i_t)$ and the importance weight is $\varepsilon(x^i_t,y^i_t)$.
\begin{equation}
{\int}\varepsilon(x^i_t,y^i_t)p(x^i_t,y^i_t)dx^i_tdy^i_t\approx \frac{1}{N_m} \sum^{N_m}_{l=1}\varepsilon(x^{i,[l]}_t,y^{i,[l]}_t),
\end{equation}
where $(x^{i,[l]}_t,y^{i,[l]}_t), l=1,2...N_m$ is the set of samples drawn from the distribution $p(x^i_t,y^i_t)$, and $N_m$ is the size of the sample set.
\indent The pseudo code of the proposed RBPF is shown as follows and depicted in Fig.1:\\
  $(C^{[k]}_t,X^{[k]}_t,w^{[k]}_t)=RBPF(C^{[k]}_{t-1},C^{[k]}_{t-1},C^{[k]}_{t-1})$
\begin{enumerate}
 \item Predict $C_t$ and $X_t$ according to Eq. (3,5)\\
 for vehicle $i=1:Nv$
 \item Determine the indicator variable according to Eq. (17)
 \item Calculate particle weights and update $X_t$ according to Eq. (19)
 \item Modify particle weights according to Eq. (24)
 \item Resample\\
  end RBPF
 \end{enumerate}

\begin{figure}[htbp]
  \centering
  \includegraphics[width=3.5in]{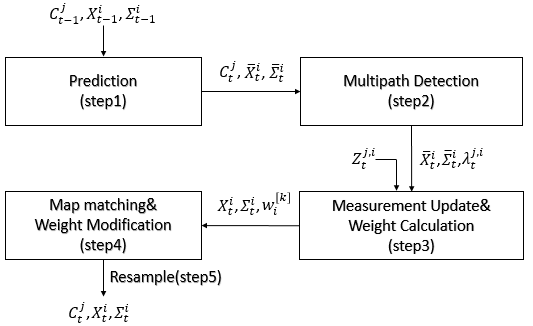}   
  \caption{Schematics of the RBPF algorithm}
\end{figure}
\section{Simulation Results and Discussions}
In this section, a simulation scenario and results are presented and discussed.
\subsection{Simulation Scenario and Noise Models}
The configuration of the simulated scenario is shown in Fig. 2, where four vehicles are traveling in each lane of two orthogonal roads, respectively. The performance of the proposed algorithm is illustrated through comparison with the CMM algorithm proposed in Rohani et al. \cite{r1}. In their approach, the common correction of the vehicle positions is searched in the position space using a particle-based approach by applying map constraints. Due to the uncertainty caused by the non-common errors, the map constraints are blurred to avoid over-convergence.\\
\begin{figure}[htbp]
  \centering
  \includegraphics[width=2.0in,scale=0.5]{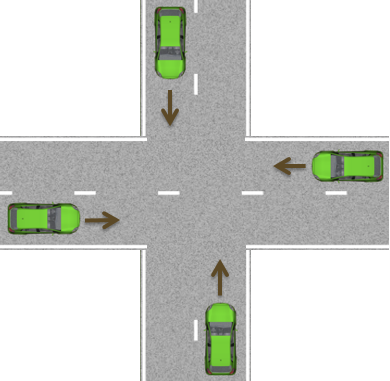}
  \caption{Configuration of the vehicles for CMM, lane width 3.5m}
\end{figure}
\indent Three CMM algorithms are compared. The first algorithm is the aforementioned one proposed by Rohani et al. (referred to as the static method); the second algorithm is a modified version of the first one, where the prior estimation of each vehicle is smoothed by a Kalman filter in the position space before implementing CMM. The third algorithm is the proposed RBPF.\\
\indent The simulation parameters appear in Table 1. The simulation uses 200 particles. In the RBPF, the initial common biases are the ground truth corrupted by white noise, with variance $\sigma^2_n=0.25\mbox{ }m^2$.\\
\begin{table}[ht] 
\caption{Simulation parameters($x$ and $y$ are local coordinates where $x$ is along the lane and $y$ is transversal to the lane)} 
\centering      
\begin{tabular}{c c c c c c c}  
\hline\hline                        
$\mbox{ }$&$\Delta t$&$Ns$&$\sigma^2_c$&$\sigma^2_{ax}$&$\sigma^2_{ay}$&$\sigma^2_b$\\ [0.1ex] 
\hline                    
Value&0.1&6&0.01&1&0.01&1  \\   
Unit&$s$&/&$m^2/s^2$&$m^2/s^4$&$m^2/s^4$&$m^2/s^2$\\ [0.3ex]       
\hline     
\end{tabular}
\begin{tabular}{c c c c c c c}  
\hline\hline                        
$\mbox{ }$&$\sigma^2_d$&$\sigma^2_z$&$\alpha_1$&$\alpha_2$&$\alpha_3$&$N_m$\\ [0.1ex] 
\hline                    
Value&1&1&0.95&1&0.99&100  \\   
Unit&$m^2/s^4$&$m^2$&/&/&/&/\\ [0.3ex]       
\hline     
\end{tabular}
\end{table}
\indent The performances of these three algorithms under two measurement noise models are simulated. The first noise model simulates common biases and uncorrelated white noise with variance $\sigma^2_z$; the second noise model simulates common biases, uncorrelated white noise with variance $\sigma^2_z$ and multipath biases. Due to the rapidly varying nature of the multipath biases, they are emulated by inserting a $4\mbox{ }m$ bias to each of the pseudo range measurements with probability $p=0.25$ randomly at each time point. The common biases and satellite geometry are emulated using the GPSoft Satellite Navigation Toolbox \cite{r19}. The satellite orbits are assumed to be perfectly circular, which means the ephemris error, as part of the common biases, is not emulated. This assumption is generally acceptable, as it forms only a small portion of the common biases. The emulated common biases include slowly varying ionospheric and tropospheric effects based on physical models of the atmosphere, which should be valid under mild atmospheric conditions.    
\subsection{Localization Results}
The horizontal position errors and associated covariance of one of the vehicles using the three described algorithms are shown in Figs. 3-6. \\
\indent Fig. 3 and Fig. 5 show that the localization error using the static method is much larger and noisier than either that using the smoothed static method or the proposed RBPF. This is to be expected, as the white noise and multipath biases in the pseudo range measurements cause significant non-common ego localization error if they are not filtered. For the first two algorithms to perform well, the non-common error must be much less than the common error. Thus, the modified algorithm significantly outperforms the static method because the non-common error caused by the white noise is effectively filtered out by the additional Kalman filter. Nevertheless, the RBPF outperforms both of them.
\begin{figure}[htbp]
  \centering
  \includegraphics[width=3.5in]{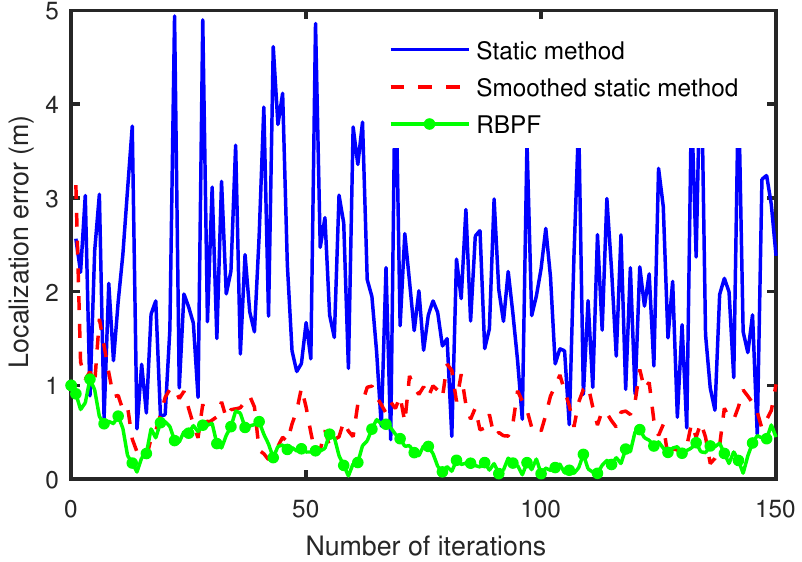}
  \caption{Horizontal position errors, noise model: common biases+white noise}
\end{figure}
\begin{figure}[htbp]
  \centering
  \includegraphics[width=3.5in]{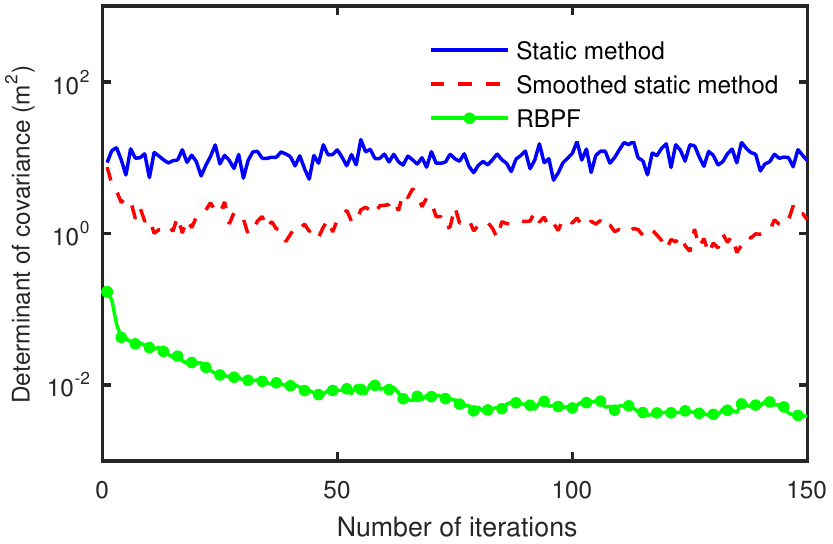}
  \caption{Determinant of horizontal position covariance, noise model: common biases+white noise}
\end{figure}
\begin{figure}[htbp]
  \centering
  \includegraphics[width=3.5in]{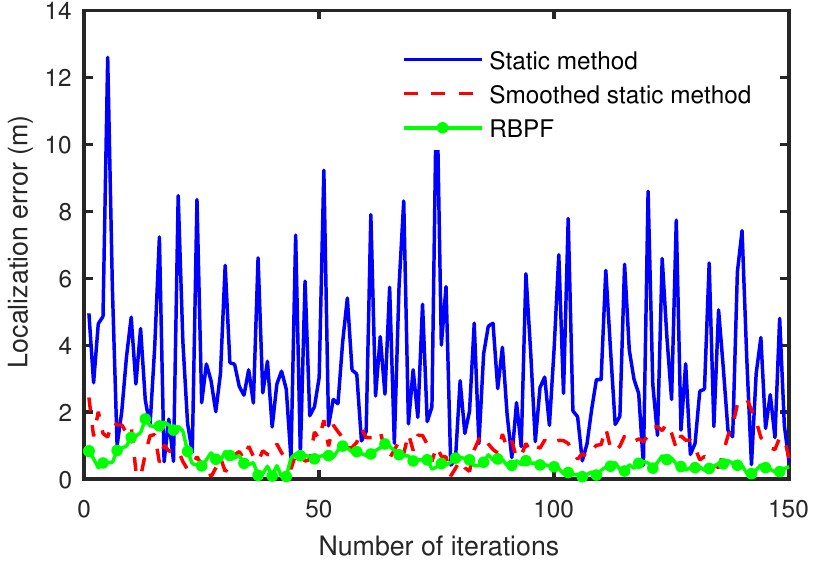}
  \caption{Horizontal position errors, noise model: common biases+white noise+multipath biases}
\end{figure}
\begin{figure}[htbp]
  \centering
  \includegraphics[width=3.5in]{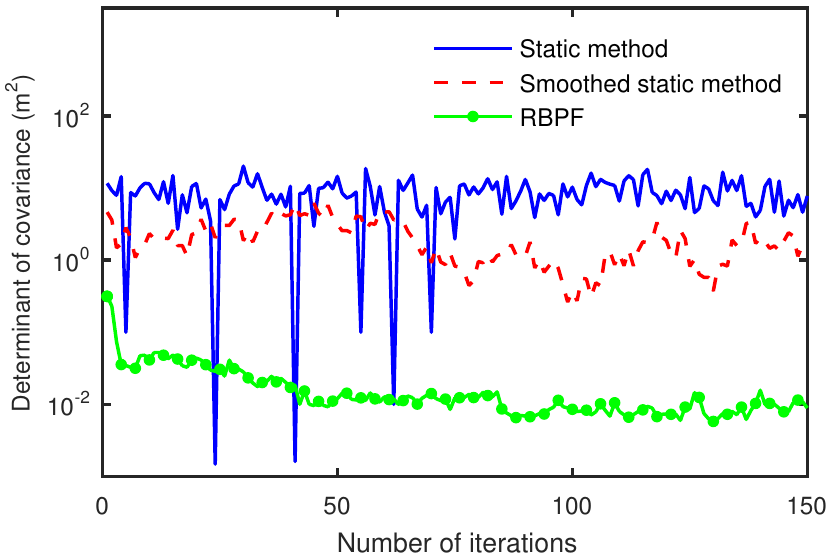}
  \caption{Determinant of horizontal position covariance: common biases+white noise+multipath biases}
\end{figure}
\indent Another benefit of the proposed RBPF is that it significantly outperforms the other two algorithms with respect to estimation effectiveness (see Figs. 4 and 6), because it does not have to search the position space exhaustively at each time step for all compatible corrections of the common error. This calculation is both unnecessary and ineffective because most of the corrections, though compatible with the current map constraints, would be eliminated if the previous localization results were also considered. In other words, the temporal correlation of the common biases is not exploited. In contrast, the proposed RBPF keeps track of the most probable common biases. The common biases of small probability are eliminated through resampling, and the time correlation of the common biases is taken advantage of by Eq. (3), that is, enforcing the estimation of the common biases to be time-continuous. Thus, the estimation covariance turns out to be much smaller than that of the first two algorithms.\\
\indent It should be noted that the estimation results from RBPF are not only effective but also consistent because the correlation between vehicle positions and the common biases are implicitly modeled by the conditionally independent filter structure. Fig.6 shows that the covariance using the static method occasionally drops below that using the RBPF, because the multipath biases induce large non-common errors, making just about all corrections of the common error incompatible. As a result, only a small number of corrections remain after resampling and the estimation result is over-confident.\\
\indent Table.2 presents the Root Mean Square (RMS) of the localization errors shown in Figs. 3 and 5.
\begin{table}[ht] 
\caption{Root Mean Square of the localization errors (m)} 
\centering      
\begin{tabular}{c c c c}  
\hline\hline                        
$\mbox{ }$ &Static &Smoothed static&RBPF\\ [0.3ex] 
\hline                    
Multipath free & 2.37 & 0.80 &0.40  \\   
Multipath included & 4.09 & 1.11&0.68\\ [0.3ex]       
\hline     
\end{tabular}
\end{table}
\subsection{Estimation Results of Common Biases}
\begin{figure*}[htbp]
\centering
\includegraphics[width=3.3in]{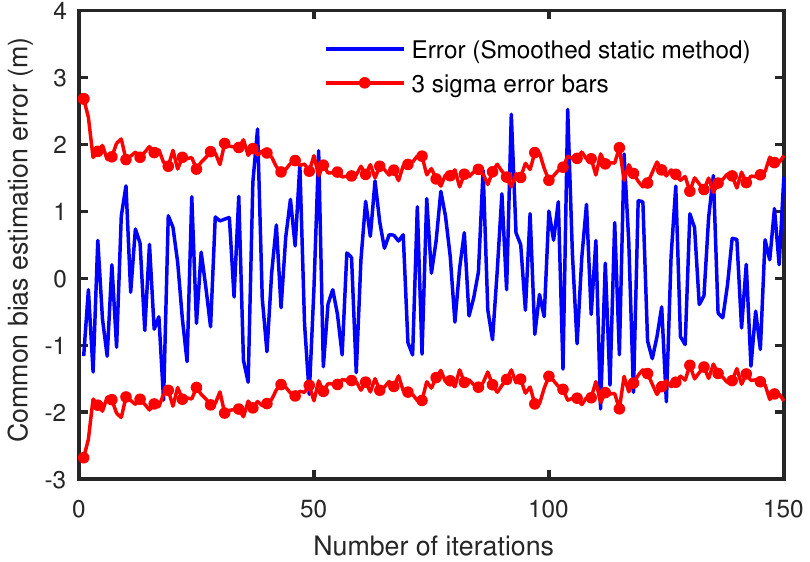}
\includegraphics[width=3.3in]{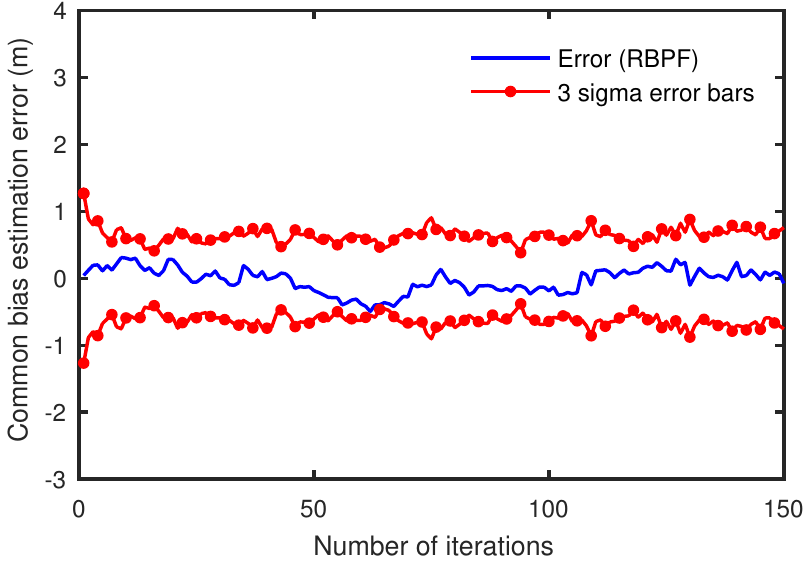}
\caption{Common bias estimation error using the smoothed static method and the associated  error bars, noise model:common biases+white noise; left: Smoothed static method, right: RBPF}
\end{figure*}
The left and right figures in Fig. 7show the common bias estimation errors corresponding to one of the satellites using the smoothed static method and the proposed RBPF and the corresponding error bars, respectively.\\
\indent The two algorithms show consistent estimation results, while the estimation from the RBPF is more accurate and effective than that using the smoothed static method. In addition, since the RBPF estimates the common biases directly while the smoothed static method uses the noisy pseudo range measurements, the estimation using the latter is much noisier than that using the former.
\section{Conclusions}
In this paper, a Rao-Blackwellized particle filter has been proposed for the simultaneous estimation of GNSS common biases and vehicles cooperative localization using map matching. The following conclusions can be drawn based on the simulation results:
\begin{enumerate}
 \item The proposed method fully exploits the temporal correlation of the common biases and vehicle positions through the prediction-update process such that the estimation covariance is reduced by at least two orders compared with previous algorithms while preserving consistency. 
 \item The proposed method effectively filters out white noise using the conditionally independent EKFs and detects and rejects measurements corrupted by multipath biases, thus significantly mitigating the effect of the non-common errors and improving  localization accuracy by about 50\% compared with previous algorithms.
 \item The proposed method estimates common biases and vehicle positions jointly. The variances of the estimated common biases are reduced by about 75\% compared with  previous algorithms.
 \end{enumerate}
\indent 
\indent For future work, the geometric dependence of the multipath biases will be investigated and exploited for better localization results.

\section*{Acknowledgment}
This work is funded by the Mobility Transformation Center at the University of Michigan.

\end{document}